\documentclass[11pt]{article}
\usepackage{coling2020}
\usepackage{times}
\usepackage{url}
\usepackage{latexsym}

\usepackage{graphicx}
\usepackage{subfigure}
\usepackage{booktabs} 
\usepackage{array}
\usepackage{caption} 
\usepackage{color} 
\usepackage{multirow}
\usepackage[labelfont=bf]{caption}

\usepackage{amsmath}
\usepackage{caption} 
\usepackage{makecell}
\usepackage{enumitem}
\colingfinalcopy 

\title{Extreme Model Compression for On-device \\Natural Language Understanding}

\author{Kanthashree Mysore Sathyendra \\
  Amazon Alexa \\
  {\tt ksathyen@amazon.com} \\\And
  Samridhi Choudhary \\
  Amazon Alexa\\
  {\tt samridhc@amazon.com} \\\And
  Leah Nicolich-Henkin \\
  Amazon Alexa\\
  {\tt nicolich@amazon.com} \\}

\date{}

\begin{document}
\maketitle
\begin{abstract}
In this paper, we propose and experiment with techniques for extreme compression of neural natural language understanding (NLU) models, making them suitable for execution on resource-constrained devices. We propose a task-aware, end-to-end compression approach that performs word-embedding compression jointly with NLU task learning. We show our results on a large-scale, commercial NLU system trained on a varied set of intents with huge vocabulary sizes. Our approach outperforms a range of baselines and achieves a compression rate of 97.4\% with less than 3.7\% degradation in predictive performance. Our analysis indicates that the signal from the downstream task is important for effective compression with minimal degradation in performance.
\end{abstract}

\section{Introduction}
\blfootnote{
%
%
 \hspace{-0.65cm} 
 This work is licensed under a Creative Commons
 Attribution 4.0 International License.
 License details:
 \url{http://creativecommons.org/licenses/by/4.0/}.
}

Spoken Language Understanding (SLU) is the task of extracting meaning from a spoken utterance. A typical approach to SLU consists of two modules: an automatic speech recognition (ASR) module that transcribes the audio into a text transcript, followed by a Natural Language Understanding (NLU) module that predicts the semantics (domain, intent and slots) from the ASR transcript. The last few years have seen an increasing application of deep learning approaches to both ASR \cite{mohamed2011acoustic,hinton2012deep,graves2013speech,bahdanau2016end} and NLU \cite{xu2014contextual,yao2013recurrent,ravuri2015recurrent,sarikaya2014application}, making them more reliable, accurate and efficient. This has led to an increasing popularity of feature-rich commercial voice assistants (VAs) \--- like Amazon Alexa, Google Assistant, Apple's Siri and Microsoft's Cortana. VAs were used in over 3 billion devices in the world in 2019, and are estimated to reach 8 billion devices by 2023\footnote{\scriptsize \url{https://www.statista.com/statistics/973815/worldwide-digital-voice-assistant-in-use/}}. With a growing number of users relying on VAs for their day-to-day activities, voice interfaces have become ubiquitous, and are employed in a range of devices, including smart TVs, mobile phones, smart appliances, home assistants and wearable devices.

The SLU processing for VAs is often offloaded to the cloud, where high-performance, compute-rich hardware is used to serve complex machine learning models. However, on-device SLU is growing in popularity due to its wide applicability and attractive benefits \cite{coucke2018snips,mcgraw2016personalized,saade2018spoken}. First, it enables VAs to work offline, without an active internet connection, allowing their use in remote areas and on devices with poor or intermittent internet connectivity, for eg. in automobiles. Second, on-device processing reduces latency by eliminating communication over the network, and results in an improved user experience. And third, processing utterances on the edge decreases the load on cloud-services, resulting in reduced cloud hardware requirements and associated costs. 

NLU is the task of extracting intents and semantics from user queries. NLU in VAs typically consists of the following sub-tasks - domain classification (DC), intent classification (IC) and named entity recognition (NER). Prior work has shown the effectiveness of recurrent neural models, that jointly model these tasks in a multi-task setup \cite{kim2017onenet,hakkani-tr2016multi-domain,Liu_2016}. These models typically are made up of large word embeddings, sometimes accounting for more than 90\% of the model parameters, and hence require compression for their deployment on resource constrained devices.  Generic model compression approaches such as quantization~\cite{hubara2017quantized} are ineffective for compressing large word-embeddings, as they do not achieve the required performance at high compression rates. Prior approaches for word-embedding compression~\cite{raunak2017simple,shu2017compressing} tackle comparatively smaller vocabulary sizes and are typically post-processing approaches, where compression is performed after the downstream task models are trained. Post-processing compression for large vocabulary sizes is not effective as the compression is lossy and task-agnostic. Under higher compression rates, post-processing word embedding compression can lead to a significant degradation in downstream performance.

In this paper, we present a principled approach for compressing neural models targeted to perform NLU on resource-constrained devices. We tackle a large number of intents and huge vocabularies ($\sim 200K$), which are typical in a large-scale, commercial NLU system. To overcome the limitations of prior task-agnostic embedding compression approaches, we propose an end-to-end compression technique, where the compression layers are jointly trained with the downstream task (NLU) model. Joint training allows for both \textit{task-aware compression} and \textit{compression-aware task learning}. \textit{Task-aware compression} enables the compression model to learn better reconstructions for words that are more important to the downstream task. At the same time, \textit{compression-aware task learning} enables the downstream task model to adapt itself to the errors in embedding reconstructions. We further combine word embedding compression with recurrent layer compression using quantization to compress our model to just a few MB, achieving a compression rate \textgreater 97\% with \textless 4\% drop in predictive performance. 

\label{intro}

\section{Related Work} \label{related work}

\textbf{Joint Modeling and Multi-Tasking for NLU:} Joint modeling of component NLU tasks, such as IC and NER, has been an extensive area of research. \newcite{jeong2008triangular} propose a triangular conditional random field (CRF) as a unified probabilistic model combining IC and NER. This is further extended by \newcite{xu2013convolutional}, where convolutional neural network based triangular CRFs are used. Other neural network architectures like recursive neural networks (RNNs) \cite{guo2014joint} and their variants \cite{zhang_2016,Liu_2016,hakkani-tr2016multi-domain,liu2016joint} have also been well explored. However, all these approaches propose to build domain specific models and produce multiple models, one for each domain. Work by \newcite{kim2017onenet} explores a unified, multi-domain, multi-task neural model using RNNs (MT-RNN) and was shown to be effective in sharing knowlege across the component tasks and domains. In contrast, the authors in \cite{hakkani-tr2016multi-domain} use a sequence-to-sequence model to output the complete semantic interpretation of an utterance (DC, IC, NER). In our work, we adapt the multi-task architecture from \newcite{kim2017onenet}, and demonstrate its effectiveness in meeting strict device constraints on compression.\\\vspace{-6pt}

\noindent \textbf{Neural Model Compression:} Due to its many practical applications, research on neural model compression has received massive interest in recent years. Existing approaches for general neural model compression include low-precision computation \cite{vanhoucke2011improving,hwang2014fixed,anwar2015fixed}, quantization \cite{chen2015compressing,zhou2017incremental}, network pruning \cite{wen2016learning,han2015learning}, SVD-based weight matrix decomposition \cite{xue2013restructuring} and knowledge distillation \cite{hinton2015distilling}. 
For neural NLP models, however, larger focus has been on compressing huge word embedding matrices. Embedding compression approaches include quantization \cite{hubara2017quantized}, binarization \cite{tissier2019near}, dimensionality reduction and matrix factorization methods such as PCA~\cite{raunak2017simple} and SVD \cite{acharya2019online}. An alternative post-training compression approach using deep compositional code learning (DCCL) was also proposed by \newcite{shu2017compressing}. This approach learns compressed embedding representations based on  additive quantization \cite{babenko2014additive} and forms the  basis of our task-aware compression approach. In contrast to \newcite{shu2017compressing}, we propose a task-aware compression approach, where embedding compression is performed during the task model training, instead of as a post-processing step.
 \pagebreak

\section{Method}

\textbf{Problem Setup:} NLU consists of three component tasks - Domain Classification (DC), Intent Classification (IC) and Named Entity Recognition (NER). DC and IC are sentence classification tasks and determine the domain (e.g. Music) and the intent (e.g. PlayMusic) of the input utterance. NER is a sequence tagging task, where each word in the utterance is assigned a slot tag (e.g. AlbumName, SongName etc). The combination of the domain, intent and slots represents the semantic interpretation for the given utterance and is passed on to the downstream application. Our goal is to compress the NLU models, to fit within extreme disk space constraints with minimal degradation in predictive performance. Furthermore, low-latency and inference support for the models are desirable.

\subsection{NLU Task Model Architecture}
\label{sec:arch}
 \begin{figure*}[tb]
 	\centering
 	\includegraphics[width=0.95\linewidth]{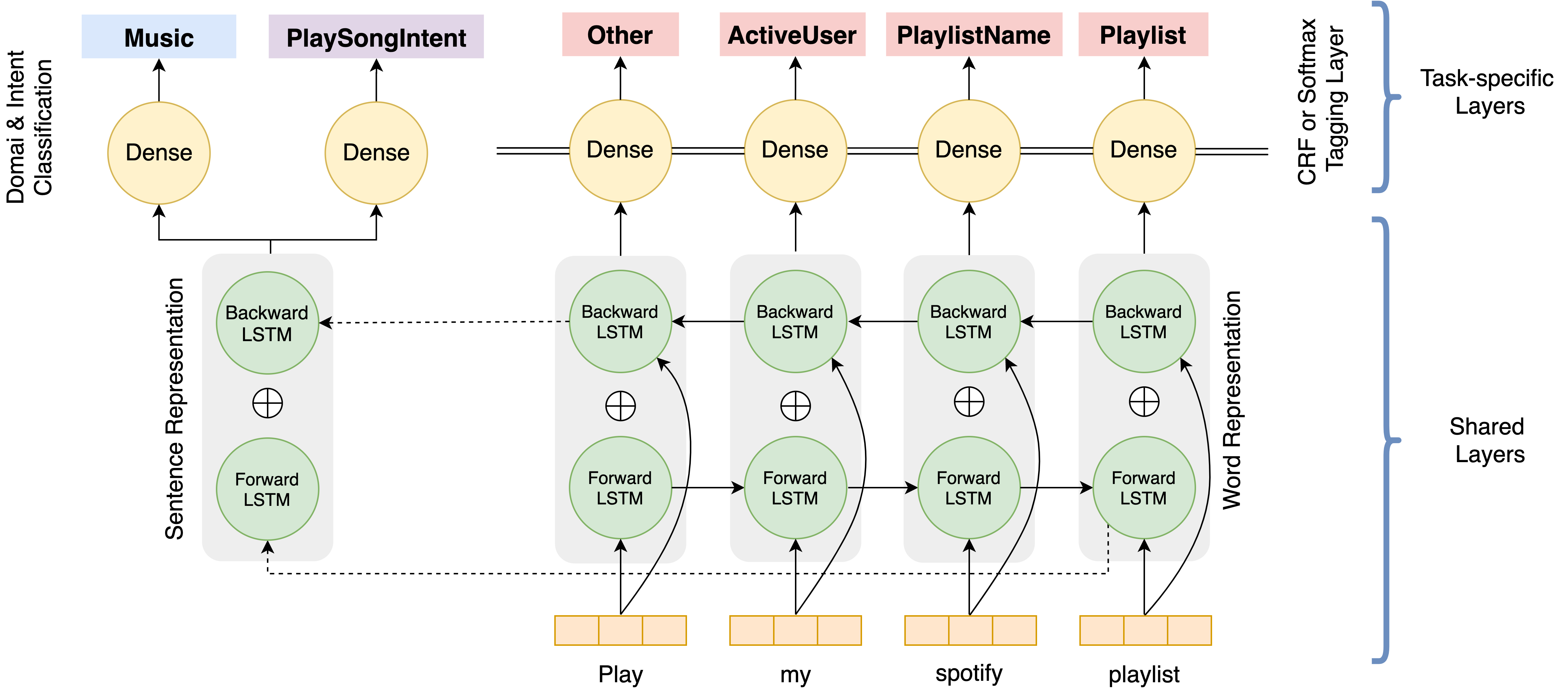}
 	\caption{Multi-Domain, Multi-Task Recurrent Architecture for on-device NLU.}
 	\label{fig:mt_model}
 \end{figure*}

\noindent \textbf{Model Architectural Constraints:} Our choice of a suitable on-device NLU architecture is largely driven by hardware resource constraints. First, on-device systems come with a strict memory budget, restricting our choices to architectures with fewer parameters. Second, the architectures chosen should not only be amenable to model compression, but should result in \textit{minimal} degradation in performance on compression. Third, on-device models have rigorous latency targets, requiring fast inference. This restricts our choices to simpler, seasoned architectures, like LSTMs and GRUs, that require fewer layers and FLOPs as opposed to the newer computationally intensive transformer-based architectures like BERT. Moreover, on-device inference engines often lack support for sophisticated layers such as self-attention layers. Driven by these constraints and relying on the considerable effectiveness of recurrent architectures \cite{hakkani-tr2016multi-domain,Liu_2016,zhang_2016}, we use a multi-domain, multi-task RNN model (MT-RNN), built using bi-directional LSTMs (Figure \ref{fig:mt_model}) for performing NLU. We train a single neural model that can jointly perform DC, IC and NER for a given input utterance. Furthermore, in order to reduce inference latency, we use word-level LSTMs as opposed to character or sub-word based models. \\\vspace{-8pt}

\noindent \textbf{Architecture Details} - Our task model, which we call the MT-RNN model, is shown in Figure~\ref{fig:mt_model}. It consists of a \textit{shared} bi-directional LSTM (Bi-LSTM) to extract features shared by all tasks, and \textit{task-specific} layers for the classification and tagging tasks. The input to the recurrent layers are pretrained embeddings and are fine-tuned during training. The input to each of the classification components is a sentence representation, obtained by concatenating the final states of the forward- and the backward-LSTM. This is passed on to a fully-connected dense layer with a softmax to predict the domain and intent for the utterance. The tagging layer produces a slot tag for each word in the utterance. The input at each time step consists of the forward- and backward-LSTM states for each word and the output is the slot tag. We choose the popularly used \textit{Conditional Random Fields (CRF)} layer for NER. The network is trained to minimize a joint NLU loss defined as the sum of the cross-entropy losses for IC and DC and the CRF loss for NER: 
$$ \mathcal{L}_{NLU} = \mathcal{L}_{DC} + \mathcal{L}_{IC}  + \mathcal{L}_{NER} $$ 

In the following sections, we describe our approach for compressing the word embeddings and the recurrent components of our MT-RNN model. 
\subsection{Word Embedding Compression}
\label{sec:we_compression}
Word embeddings have been shown to be the largest components in an NLP model, owing to large vocabulary sizes and floating point parameters, accounting for \textgreater 90\% of the model sizes \cite{shu2017compressing}. Hence, compressing embeddings is crucial for reducing NLP model sizes. 

Our approach is based on additive quantization~\cite{babenko2014additive}, which has shown great success in compressing word embeddings, achieving high compression rates \cite{shu2017compressing}. 

\subsubsection{Additive Quantization using Deep Compositional Code Learning} Additive quantization \cite{babenko2014additive} aims to approximate vectors by representing them as a sum of basis vectors, called codewords. Originally proposed for image compression and approximate nearest neighbor search, this method has recently been used for post-processing word embedding compression~\cite{chen2018learning,shu2017compressing} achieving high compression rates, upwards of 90\%, on modest vocabulary sizes. 

Let $W \in \mathcal{R}^{V \times D}$ be the original word embedding matrix, where $V$ denotes the vocabulary size and $D$ denotes the embedding size. Using additive quantization, the original word embedding matrix is compressed into a matrix of integer codes as $W_c \in \mathcal{Z_K}^{V \times M}$, where $\mathcal{Z_K}$ denotes the set of integers from $1$ to $K$, $\mathcal{Z_{K}} = \{1,2,\ldots,K\}$. This is achieved using a set of $M$ codebooks, $C_1$ through $C_M$, $C_m \in \mathcal{R}^{K \times D}$, each containing $K$ codewords of size $D$. $C_m^k$ is the $k^{\text{th}}$ codeword in the $m^{\text{th}}$ codebook. 
For each word embedding $w_i \text{ in } W$, the compressed codes can be $w_{ci}$, where
$$w_{ci} = [z_1^i, z_2^i, \ldots, z_M^i] \qquad \text{where } z_m^i \in \mathcal{Z_K}, \forall m \in \{1,2,\ldots,M\} $$
 
The original word embedding $w_i$ is approximated from the codes and codebooks as $w_i'$ by summing the $(z_m^i)^{\text{th}}$ codeword in the $m^{\text{th}}$ codebook over all codebooks: $$ w_i' = \sum_{m=1}^M C_m^{z_m^i}$$

 \begin{figure*}[tb]
 	\centering
 	\includegraphics[width=1\linewidth]{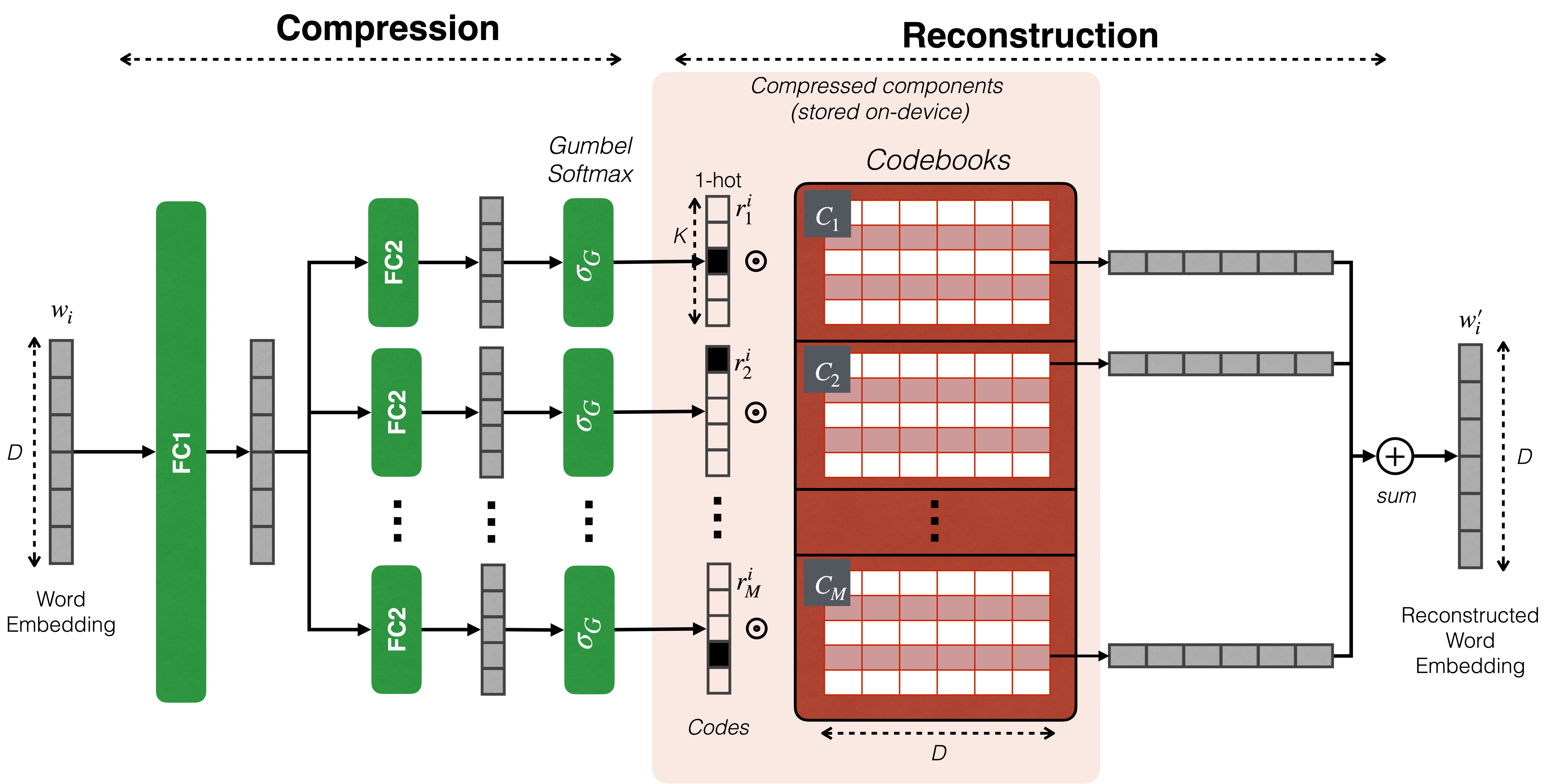}
 	\caption{Deep Compositional Code Learning Architecture.}
 	\label{fig:code_learning}
 \end{figure*}

\newcite{shu2017compressing} propose the deep compositional code learning (DCCL) architecture to learn discrete codes and codebooks for a given word embedding matrix through an unsupervised autoencoding task. In this model, a continuous word vector input, $w_i \in \mathcal{R}^{D}$ is first projected into a lower dimensional space using a linear transformation. This is projected through a second linear layer into $M$ different $K$-dimensional vectors. Each of these $M$ vectors is passed through a gumbel-softmax activation to get $M$ one-hot vectors, $r_m^i \in \mathcal{R}^{1 \times K}$:
$$ r_m^i = \sigma_G(f_L({w_i})) \qquad \forall m \in \{1,2,\ldots,M\}$$

where $f_L$ denotes the linear transformations and $\sigma_G$ denotes the gumbel-softmax activation. The gumbel-softmax activation allows the network to learn discrete codes via gumbel-sampling, while also making the network differentiable, enabling the backpropagation of gradients \cite{jang2016categorical}. 

These one-hot vectors are converted to integer codes corresponding to the input word embedding. In order to reconstruct the word embedding, the following operations are performed:

\begin{equation}
\label{reconstruct}
w_i' = \sum_{m=1}^M r_m^i*C_m \qquad \text{where } r_m^i \in \mathcal{R}^{1 \times K}, C_m \in \mathcal{R}^{K \times D}, w_i' \in \mathcal{R}^{1 \times D}
\end{equation}

Figure ~\ref{fig:code_learning} provides an overview of the DCCL model. Since the word embedding matrix $W$ can be reconstructed using just the codes $W_c$ and the codebooks $C=[C_i\ldots C_m]$, the original embedding matrix $W$ with $V \times D$ floating point values need not be stored on-device, thus achieving the required compression. Furthermore, $W_c$ would be an integer matrix requiring only $M\log_2 K$ bits per embedding and the codebook $C$ requires just $M*K*D*32$ bits on disk, where each floating point element takes 32 bits. By choosing $M$ and $K \ll V$, the size of the codes and codebooks can be greatly reduced when compared to the original embedding matrix.

\subsubsection{Task-agnostic Post-Processing Compression}
\newcite{shu2017compressing} propose to use the DCCL architecture to perform post-processing embedding compression, where embeddings are compressed after the downstream task model has been trained. The task model is first initialized with pretrained word embeddings that are fine-tuned during task model training to obtain task-specific embeddings. These are compressed using the DCCL architecture trained on an unsupervised autoencoding task. The input to the autoencoder is the embedding matrix $W \in \mathcal{R}^{V \times D}$ and the model is trained to minimize the average embedding reconstruction loss (denoted by $l(W, W')$) for words in the embedding matrix:  
$$l(W, W') = \frac{1}{V}\sum_{i=1}^V (w_i - w_i')^2$$
DCCL is shown to outperform other approaches such as parameter pruning and product quantization on sentence classification and machine translation tasks. 

Since compression is performed as a post-processing step after the task model is trained, the compression algorithm has no information about the downstream task, making the compression task-agnostic and results in several drawbacks. First, unsupervised post-processing compression treats all words equally for compression. However, in practice, some words may be more important than others for the downstream task. Hence, better reconstructions of more important words may benefit the downstream task. Second, post-processing compression typically is lossy resulting in a degradation in downstream performance since the task model is not adapted to the compression error. We propose a task-aware end-to-end compression approach which aims to address these issues. 

 \pagebreak
\subsubsection{Task-aware End-to-End Compression}
\label{sec:ta_comp}
 \begin{figure*}[tb]
 	\centering
 	\includegraphics[width=1\linewidth]{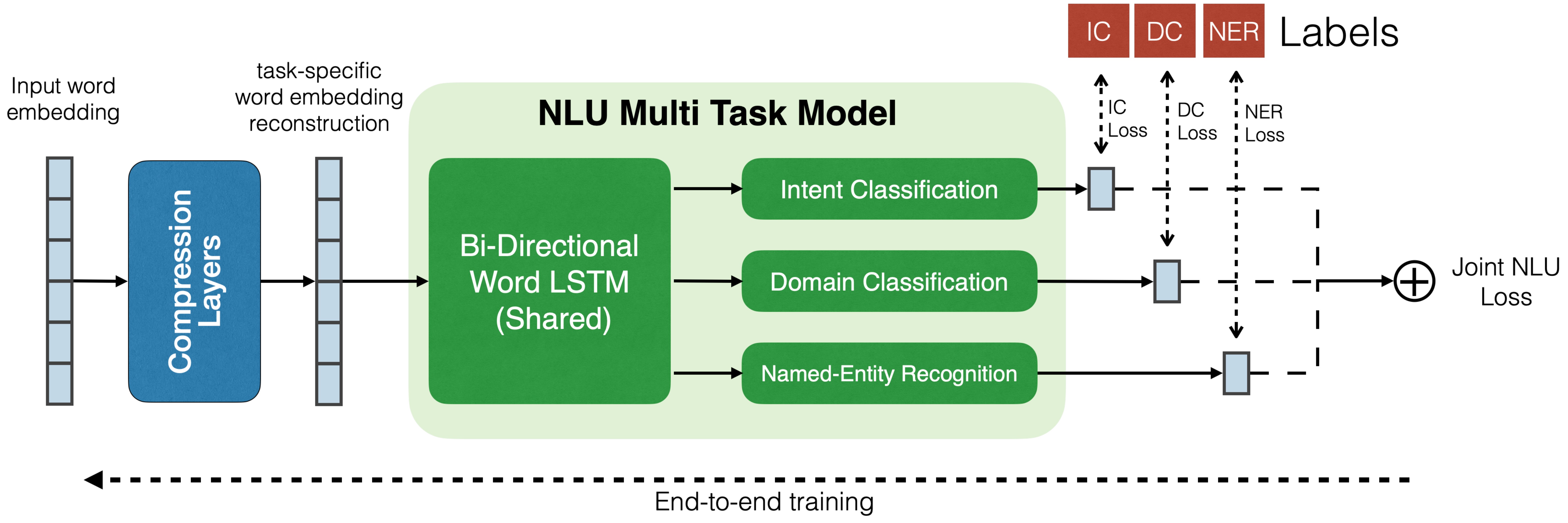}
 	\caption{Task-aware end-to-end compression with the MT-RNN model.}
 	\label{fig:e2e}
 \end{figure*} 
Our algorithm improves on the above said approach, by training the DCCL a.k.a. the compression model, jointly with the downstream task model (Figure \ref{fig:e2e}). End-to-end training allows the compression model to receive signals about the downstream task, thus adapting the compression to the downstream task. Intuitively, since the compression model now has the information about how the words are used in the downstream task (via the downstream loss), it can spend more network capacity in achieving better reconstructions for more important words. At the same time, the downstream task model also adapts to the lossy reconstructions learned by the compression model, thus improving on the downstream performance. We call this \textit{task-aware end-to-end compression}, where the compression algorithm takes the downstream task loss into account during embedding compression. 

In order to perform task-aware compression with a DCCL model, we replace the original embedding lookup operations in the task model with layers from the DCCL model a.k.a. the compression layers. The input to our model is now a sequence of $L$ word embeddings corresponding to words from the input text utterances. These are passed through the compression layers and are reconstructed, as shown in equation \ref{reconstruct}, to obtain a sequence of $D$ dimensional word representations corresponding to each word in the input. The word representation is then fed to the recurrent layers in the task model and the remaining network is unchanged. The entire setup is trained end-to-end to minimize the downstream task loss and the gradients are back-propagated through the entire network, including the compression layers. Further, the compression layers can be initialized with pretrained model parameters from the task-agnostic DCCL model, and the NLU layers can be initialized from a trained NLU model. 

Training an end-to-end DCCL model is tricky, especially when the number and size of codebooks is large. The stochasticity introduced by gumbel-sampling can easily stray off the training, leading to sub-optimal convergence. For these cases, we ground the training by adding 
the word embedding reconstruction loss to the downstream task loss as follows: 

\begin{equation*}
\mathcal{L} = \mathcal{L}_{NLU} +  \mathcal{L}_{e} \text{ where }  \mathcal{L}_{e}  = \frac{1}{N}\sum_{i=1}^N \Big(w_i - w_i'\Big)^2 
\vspace{-5pt}
\end{equation*}
Adding the embedding reconstruction loss not only stabilizes the training, but also provides stronger gradients to the compression layers. Note that unlike task-agnostic compression where all words are treated equally for compression, the embedding reconstruction loss term in task-aware compression considers only the words appearing the in the input batch. This ensures that the words that are more frequent in the training data have better reconstructions, resulting in better downstream performance.

\subsection{Recurrent Layer Compression} Quantization~\cite{hubara2017quantized} is a simple and effective technique for model compression. Quantization maps each floating point model paramater to it’s closest representative from a pre-chosen set of floating-point values. More concretely, the model parameter range is divided into $B$ equally spaced bins (or buckets), and each parameter is assigned it’s closest bin. The bins can be represented by integer indices and require at most $\log_2 B$ bits. For instance, with 256 bins, a 32-bit floating point parameter can represented by an integer bin index occupying just 8 bits. 

We apply post-training 8-bit linear quantization to quantize the recurrent layers of the model. Since 32-bit floating point model parameters are now represented by 8-bit integers, this results in an instant 4$\times$ compression. Furthermore, quantization improves model latency, as all the floating point operations are performed using integers. While more sophisticated compression techniques exist for compressing recurrent layers, we found that quantization was extremely effective and resulted in no degradation in performance. 

\section{Experiments}

\label{sec:data}
In this section we describe the datasets used and our experimental setup for model compression. While our approach is generically applicable to any NLP task that uses word embeddings, we show the effectiveness of our approach on the three NLU tasks -- DC, IC, and NER. We show our results on a large scale commercial NLU system trained across a large number of intents with huge vocabularies.\\\vspace{-6pt}

\noindent\textbf{Dataset.}
We use annotated live traffic data of a large-scale, cloud based, commercial VA system to train our NLU models. Utterances from the live traffic are randomly sampled and anonymized to remove any customer specific information. They are then annotated by skilled annotators for the NLU labels corresponding to the domain, intent and slot labels for each utterance. The training set chosen for our experiments contains millions of utterances spanning 5 domains, and over 150 intents and slots. One of these domains is the `Out of domain' (or OOD) domain, consisting of utterances not supported by the NLU system. The intent for these utterances is labeled as the `OODIntent' and the words are given the `Other' slot tag. Our held-out test set is prepared by randomly sampling 1 million utterances from the live-production traffic, following a similar process. In order to facilitate optimization and early stopping, we also use a validation set of a similar scale.\\\vspace{-6pt}

\noindent\textbf{Evaluation Metrics.}
We use the following metrics for evaluating the performance on the NLU tasks:

\textbf{\textit{Intent Recognition Error Rate (IRER)}}: This is the ratio of number of incorrect interpretations to the total number of utterances. A correct interpretation is when the predicted domain, intent and all slots for an utterances are correct. We compute the \textit{IRER} only on non-OOD utterances.

\textbf{\textit{Intent Classification Error Rate (ICER)}}: This is the ratio of number of incorrect intent predictions to the total number of utterances.

\textbf{\textit{Domain Classification Error Rate (DCER)}}: This is the ratio of number of incorrect domain predictions to the total number of utterances.

\textbf{\textit{Slot Error Rate (SER)}}: This is the ratio of number of incorrect slot predictions to the total number of slots.

\textbf{\textit{False Accept Rate (FAR)}}: This is the ratio of number of out-of-domain utterances falsely accepted as a supported utterance to the total number of out-of-domain utterances. This metric is mainly used to evaluate the effectiveness of the model in rejecting out-of-domain (or unsupported) utterances.

Along with the above metrics we also compute the sizes of the word embeddings and the MT-RNN task model. We only report relative changes in the above metrics compared to the baseline.\\\vspace{-6pt}

\noindent\textbf{NLU Model Training.} 
We train the NLU task model (the MT-RNN model) described in Section \ref{sec:arch} using the prepared training dataset (Section \ref{sec:data}). We initialize the embeddings with FastText \cite{joulin2016fasttext} embeddings that have been pretrained on a large corpus of unannotated, anonymized, live utterances. The model is trained to minimize the NLU loss $\mathcal{L}_{NLU}$ as described in Section~\ref{sec:arch} and the embeddings are fine-tuned during training. The models are trained for a total of 25 epochs, with early stopping on the validation loss, using Adam optimizer with a learning rate of 0.0001. We further perform a grid search on a range of hyperparameter values for dropout and variational dropout and select the best performing model as our candidate model for compression. This model also serves as our uncompressed baseline. \\\vspace{-6pt}

\noindent\textbf{Baselines.} We compare our proposed approach with the following baselines. We use the abbreviations `TAg.' for `Task Agnostic' and `TAw.' for `Task Aware'.\\\vspace{-8pt}

\textbf{\textit{TAg. SVD}}: In this approach, large embedding matrices are factorized into matrices of much smaller sizes to produce low-rank approximations of the original embedding matrix, using Singular Value Decomposition (SVD). This is applied as an offline compression method where the embedding matrices are compressed as a post-processing step.\\\vspace{-8pt}

\textbf{\textit{TAw. SVD}}: \newcite{acharya2019online} propose a task-aware SVD-based embedding compression approach, where the embedding matrix is first factorized into lower dimensional matrices using SVD. The factors are then used to initialize a smaller word embedding layer followed by a linear layer, and jointly fine-tuned with the downstream task model. Stochastic Gradient Descent (SGD) with a learning rate of 0.001 as presented in \newcite{acharya2019online} is used for the optimizer.\\\vspace{-8pt}

\textbf{\textit{TAg. DCCL}}: Task-agnostic compression method proposed by \newcite{shu2017compressing} where the code learning autoencoder described in Section \ref{sec:we_compression} is used to compress word-embeddings from the trained NLU model. Since it does not perform joint training of the compression layers with the downstream task, this serves as an ablation test for our proposed task-aware compression approach. \\\vspace{-8pt}

\textbf{\textit{TAg. DCCL + NLU Finetuning}}: This is another ablation test for our proposed task-aware compression approach. In this approach, task-agnostic compression is performed as in the previous baseline. Once compressed in a task-agnostic way, the embeddings are kept frozen and the downstream task model is fine-tuned to minimize the downstream NLU loss. NLU model fine-tuning is performed with a learning rate of 0.0001 for 5 epochs.\\\vspace{-8pt}

For all SVD-based approaches, we run experiments over a range of values for $n$ where $n$ is the fraction of components retrained in the low-rank SVD approximation. This produces models of different sizes. For all DCCL-based baselines, we train the task-agnostic autoencoder model for 300 epochs (approximately 800k iterations) with a learning rate of 0.0001 using the Adam optimizer. We experiment with a range of values for hyperparameters $M$ and $K$ where $M$ is the number of codebooks and $K$ is the number of basis vectors per codebook. Different values of $M$ and $K$ produce models of different sizes.\\\vspace{-6pt}

\setlength{\tabcolsep}{4pt}
\begin{figure}[!tb]
    \centering
    \begin{minipage}{0.61\textwidth}
        \centering
       {\small
\begin{tabular}{@{}llcccc@{}}
\toprule
\multirow{2}{*}{Model}           & \multicolumn{1}{c}{\multirow{2}{*}{Type}} & \multicolumn{4}{c}{Word Embedding Compression Rate}               \\
                                 & \multicolumn{1}{c}{}                      & 15$\times$             & 30$\times$             & 60$\times$             & 120$\times$            \\ \midrule
SVD                              & TAg.                                      & +786.94        & +794.11        & +794.43        & +794.42        \\
DCCL                             & TAg.                                      & +12.14         & +81.67         & +280.89        & +530.54        \\
DCCL + NLU Fine-tuning           & TAg.                                      & +1.49          & +3.78          & +6.13          & +8.84          \\
SVD                              & TAw.                                      & +82.20         & +106.06        & +161.70        & +235.68        \\ \midrule
Ours                             & TAw.                                      & \textbf{+0.84} & \textbf{+2.45} & +4.87          & +8.85          \\
Ours -- w.o. recons. loss & TAw.                                      & +5.91          & +3.79          & \textbf{+3.72} & \textbf{+5.53} \\
Ours -- w.o. pretraining         & TAw.                                      & +7.23          & +5.39          & +6.30          & +8.71          \\ \bottomrule
\end{tabular}
\captionof{table}{Relative percentage IRER change for different word embedding compression rates.}
\label{tab:comp}
}
    \end{minipage}
    \hfill
    \begin{minipage}{0.35\textwidth}
        \centering
        \includegraphics[width=0.91\linewidth]{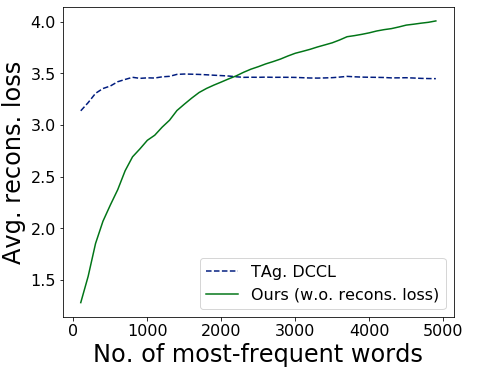}
        \vspace{-7pt}
        \caption{Average Reconstruction loss for top frequent words.}
        \label{fig:recons_loss}
    \end{minipage}%
\end{figure}

\noindent\textbf{Implementation details.} Our approach is essentially a task-aware version of DCCL (\textit{\textbf{TAw. DCCL}}). In our method, the compression layers are initialized with the parameters from the trained autoencoder model, obtained as a result of task-agnostic post-processing compression. Similarly, the NLU specific layers are initialized from the trained NLU model. The entire compression model is then trained end-to-end to minimize the loss function as mentioned in Section~\ref{sec:ta_comp}. The model is trained with a learning rate of 0.0001 for 5 epochs. Similar to the above task-agnostic setups, we experiment with a range of values for $M$ and $K$. We further explore the following additional setups:\\\vspace{-8pt}

\textbf{\textit{Without pretraining:}} In this setup, the compression layers and the task model are jointly trained from scratch and are not initialized from pretrained components. The model is trained to minimize the joint NLU loss without the embedding reconstruction loss. We use the Adam optimizer with a learning rate of 0.0001 and train the model for 25 epochs.\\\vspace{-8pt}

\textbf{\textit{Without embedding reconstruction loss:}} In this approach, we do not add the embedding reconstruction loss to the downstream task loss. The models are, however, initialized from pretrained components, and trained end-to-end for 5 epochs.\\\vspace{-6pt}

\section{Results and Analysis}

\setlength{\tabcolsep}{4pt}
\begin{table*}[t]
\centering
{
\small
\scalebox{0.935}{
\begin{tabular}{@{}llcccccccc@{}}
\toprule
                                 & \multicolumn{3}{c}{Compression}                                                                                                      & \multicolumn{1}{l}{} & \multicolumn{5}{c}{Performance}                                                                                                                                                                                                                                                                                                              \\ \midrule
Method                           & Type & \begin{tabular}[c]{@{}c@{}}Model \\ Compr.\\ Rate\end{tabular} & \begin{tabular}[c]{@{}c@{}}WE \\ Compr.\\  Rate\end{tabular} &                      & \begin{tabular}[c]{@{}c@{}}Rel. IRER\\  Change (\%)\end{tabular} & \begin{tabular}[c]{@{}c@{}}Rel. ICER\\   Change (\%)\end{tabular} & \begin{tabular}[c]{@{}c@{}}Rel. DCER \\  Change (\%)\end{tabular} & \begin{tabular}[c]{@{}c@{}}Rel. SER \\ Change (\%)\end{tabular} & \begin{tabular}[c]{@{}c@{}}Rel. FAR\\  Change (\%)\end{tabular} \\ \midrule
Uncompressed                     & NA   & 1$\times$                                                           & 1$\times$                                                           &                      & 0                                                                & 0                                                                 & 0                                                                 & 0                                                               & 0                                                               \\ \midrule
SVD                              & TAg. & 17.4$\times$                                                           & 60$\times$                                                           &                      & +794.43                                                          & +1929.50                                                          & +2387.46                                                          & +1107.66                                                        & +2.78                                                           \\
DCCL                             & TAg. & 17.6$\times$                                                           & 64$\times$                                                           &                      & +280.89                                                          & +257.63                                                           & +282.29                                                           & +291.10                                                         & \textbf{-0.28}                                                  \\
DCCL + NLU Finetuning            & TAg. & 17.6$\times$                                                           & 64$\times$                                                           &                      & \textbf{+6.13}                                                   & \textbf{+5.53}                                                    & \textbf{+5.93}                                                    & \textbf{+7.77}                                                  & +0.06                                                           \\ \midrule
SVD                              & TAw. & 17.4$\times$                                                           & 60$\times$                                                           &                      & +161.70                                                          & +190.05                                                           & +180.76                                                           & +169.53                                                         & +0.94                                                           \\
Ours                             & TAw. & 17.6$\times$                                                           & 64$\times$                                                           &                      & +4.87                                                            & +4.50                                                             & +5.57                                                             & +6.13                                                           & \textbf{+0.07}                                                  \\
Ours -- w.o. Reconstruction Loss & TAw. & 17.6$\times$                                                           & 64$\times$                                                           &                      & \textbf{+3.72}                                                   & \textbf{+3.66}                                                    & \textbf{+4.23}                                                    & \textbf{+4.86}                                                  & +0.08                                                           \\
Ours -- w.o. Pretraining         & TAw. & 17.6$\times$                                                           & 64$\times$                                                           &                      & +6.30                                                            & +6.08                                                             & +6.70                                                             & +7.50                                                           & +0.08                                                           \\ \midrule
Ours + NLU LSTM Quantization     & TAw. & 39.5$\times$                                                           & 64$\times$                                                           &                      & \textbf{+3.69}                                                   & \textbf{+3.67}                                                    & \textbf{+4.27}                                                    & \textbf{+4.91}                                                  & \textbf{+0.08}                                                  \\ \bottomrule
\end{tabular}
}}
\caption{ This table shows relative performance metrics and model sizes for different baselines and our proposed approaches. The best models in each category are highlighed in bold.}
\label{tab:summary}
\end{table*}

Table~\ref{tab:comp} summarizes the impact of various word embedding compression approaches on the downstream IRER metric for a range of compression rates. \textit{Compression rate} is determined by dividing the uncompressed embedding (or model) size by the compressed embedding (or model) size. We report percentage relative changes\footnote{Absolute numbers are not provided due to commercial confidentiality requirements.} to the IRER when compared to the uncompressed baseline. The results presented are for 300 dimensional embeddings. However, similar trends were observed for 100 dimensional embeddings as well. 

In general, we find that task-aware approaches perform better than task-agnostic post-processing approaches. This is because the task-aware end-to-end compression tunes the compression to the downstream task, while also adjusting the task model parameters to recover performance due to lossy reconstructions. From Table~\ref{tab:comp} we also find that for any given compression rate, our proposed task-aware DCCL approach has the least degradation in predictive performance when compared to other methods. 

Task-aware DCCL outperforms even the best task-agnostic compression baseline (TAg. DCCL + NLU Fine-tuning) by 39-44\% at each of the different compression rates. This shows that the loss signal from the downstream task helps performance by not only adapting the task model to the compression, but also by improving compression quality. Moreover, our model at 120$\times$ compression rate performs better than the best baseline even at 60$\times$ compression rate. In other words, our models are 2$\times$ smaller than even the best baseline for a similar performance. We also find that the embedding reconstruction loss added to the downstream task loss helps improve the downstream performance, especially when the compression rate is lower i.e. when the gumbel-sampling layers are larger or more in number. 

In order to understand the importance of task-aware compression, we plot the word embedding reconstruction loss (Figure \ref{fig:recons_loss}) for the top most frequent words in our dataset. As seen in Figure~\ref{fig:recons_loss}, the average reconstruction loss for task-agnostic DCCL remains approximately constant irrespective of frequency of the words, indicating that all words are treated equally. In contrast, task-aware compression reduces the average reconstruction loss for more frequent words indicating that the network capacity is spent to learn better reconstructions for words more important for the downstream task. Note that the model used for the graph is the task-aware DCCL model without the reconstruction loss term.

We also find that DCCL-based approaches consistently performed better than their SVD counterparts, in both task-aware and task-agnostic variants. SVD-based approaches do not perform well beyond a specific compression rate (+7.99\% for 1.7$\times$ compression). On investigating, we found that word embeddings were full rank matrices, with high singular values for all components, indicating that these components captured high variance. 

Table~\ref{tab:summary} presents a summary of the performance of the best models for each of the approaches at around 60$\times$ embedding compression rate. 8-bit Bi-LSTM quantization helps reduce the size of the recurrent layers in the models, resulting in a net model compression ratio of 39.5$\times$ with a minimal performance degradation of 3.69\% when compared to the uncompressed baseline.

\section{Conclusion}
In this paper, we present approaches for extreme model compression for performing natural language understanding on resource-constrained device. We use a unified multi-domain, multi-task neural model that performs DC, IC and NER for all supported domains. We discuss model compression approaches to compress the bulkiest components of our models - the word embeddings, and propose a task-aware end-to-end compression method based on deep compositional code learning where we jointly train the compression layers with the downstream task. This approach reduced word embeddings sizes to just a few MB, achieving a word-embedding compression rate of 98.4\% and outperforms all other task-agnostic and task-aware embedding compression baselines. We further apply post-training 8-bit linear quantization to compress the recurrent layers of the model. These approaches together result in a net model compression rate of 97.5\%, with a minimal performance degradation of 3.64\% when compared to the uncompressed model baseline. 

DCCL approaches are complementary to other compression approaches such as knowledge distillation and model pruning. While our work demonstrates the effectiveness of task-aware DCCL on the classification and tagging tasks in NLU, the approach itself is generic and can be applied to other NLP tasks that rely on large word-embeddings. As part of future work, we would like to explore the effectiveness of task-aware DCCL on NLP tasks such as machine translation and language modeling. We would also like to explore compression of models with advanced architectures using contextual embeddings.

\bibliographystyle{coling}
\bibliography{arxiv_2020}

\end{document}